\documentclass{article}

\usepackage{arxiv}
\usepackage[utf8]{inputenc} % allow utf-8 input
\usepackage[T1]{fontenc}    % use 8-bit T1 fonts
\usepackage{hyperref}       % hyperlinks
\usepackage{url}            % simple URL typesetting
\usepackage{booktabs}       % professional-quality tables
\usepackage{amsfonts}       % blackboard math symbols
\usepackage{nicefrac}       % compact symbols for 1/2, etc.
\usepackage{microtype}      % microtypography
\usepackage{cleveref}       % smart cross-referencing
\usepackage{float} 
\usepackage{graphicx}
\usepackage{natbib}
\usepackage{doi}

\title{ReflectivePrompt: Reflective evolution in autoprompting algorithms}

\author{ \href{https://orcid.org/0009-0009-0788-9790}{\includegraphics[scale=0.06]{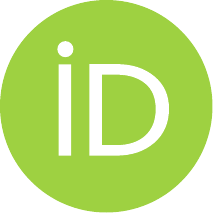}\hspace{1mm}Viktor N. Zhuravlev} 
\href{https://orcid.org/0009-0007-0442-7764}{\includegraphics[scale=0.06]{orcid.pdf}\hspace{1mm}Artur R. Khairullin} \href{https://orcid.org/0009-0000-3865-4446}{\includegraphics[scale=0.06]{orcid.pdf}\hspace{1mm}Ernest A. Dyagin} \href{https://orcid.org/0009-0002-6046-8943}{\includegraphics[scale=0.06]{orcid.pdf}\hspace{1mm}Alena N. Sitkina} \href{https://orcid.org/0000-0002-3952-6080}{\includegraphics[scale=0.06]{orcid.pdf}\hspace{1mm}Nikita I. Kulin} \\ 
	Computer Technologies Laboratory\\
	ITMO University\\
	Saint-Petersburg, Russia \\
	\texttt{334857@niuitmo.ru 242106@niuitmo.ru 368983@niuitmo.ru}
}

\hypersetup{
pdftitle={ReflectivePrompt: Reflective evolution in autoprompting algorithms},
pdfsubject={cs.AI, cs.CL, cs.ML},
pdfauthor={Viktor N. Zhuravlev},
pdfkeywords={AutoPrompting, LLM, NLP, Reflective Evolution},
}

\begin{document}

\makeatletter
\@ifundefined{undertitle}{\newcommand{\undertitle}{}}{}
\@ifundefined{headeright}{\newcommand{\headeright}{}}{}
\@ifundefined{shorttitle}{\newcommand{\shorttitle}{}}{}
\@ifundefined{shortauthor}{\newcommand{\shortauthor}{}}{}
\makeatother

\providecommand{\undertitle}{}
\maketitle
\begin{abstract}
Autoprompting is the process of automatically selecting optimized prompts for language models, which has been gaining popularity with the rapid advancement of prompt engineering, driven by extensive research in the field of large language models (LLMs). This paper presents ReflectivePrompt\footnote{Code available as a part of CoolPrompt framework library: https://github.com/CTLab-ITMO/CoolPrompt/} — a novel autoprompting method based on evolutionary algorithms that employs a reflective evolution approach for more precise and comprehensive search of optimal prompts. ReflectivePrompt utilizes short-term and long-term reflection operations before crossover and elitist mutation to enhance the quality of the modifications they introduce. This method allows for the accumulation of knowledge obtained throughout the evolution process and updates it at each epoch based on the current population. ReflectivePrompt was tested on 33 datasets for classification and text generation tasks using open-access large language models: t-lite-instruct-0.1 and gemma3-27b-it. The method demonstrates, on average, a significant improvement (e.g., 28\% on BBH compared to EvoPrompt) in metrics relative to current state-of-the-art approaches, thereby establishing itself as one of the most effective solutions in evolutionary algorithm-based autoprompting.
\end{abstract}

% keywords can be removed

\section{Introduction}
Large Language Models (LLMs) have demonstrated significant results in solving Natural Language Processing (NLP) tasks \cite{wei2021llms, kadavath2022llms}. Prompting and prompt engineering are universal methods for improving the performance of LLMs that do not require access to model weights and gradients during training. Instead, they enhance the efficiency of LLM inference by providing carefully crafted and well-structured instructions (prompts) as input to the model \cite{liu2021prompting}. Currently, there are many different prompting techniques, such as Few-Shot \cite{brown2020fewshot}, Role-Based \cite{wang2023rolellm}, Chain-of-Thought \cite{wei2022chainofthought}, Plan-and-Solve \cite{wang2023planandsolve}, and others. What all these techniques have in common is that they can be time-consuming to manually create, iterate, and optimize, often requiring expert knowledge and experience. The reason for this is that models are highly sensitive to input data, necessitating careful and precise application of these techniques \cite{leidinger2023languageofprompting}.

Autoprompting addresses this issue by automating the generation and selection of prompts \cite{shin2020autoprompt}. It is based on various optimization methods and principles, including reinforcement learning, evolutionary, gradient-based, and gradient-free approaches, among others \citep{shin2020autoprompt, kwon2024stableprompt, guo2023evoprompt, prasad2022grips}. In particular, prompt optimization can be either discrete or continuous \cite{schulhoff2024promptreport}. Continuous optimization involves representing the prompt as a numerical tensor, while discrete optimization treats the prompt as a sequence of tokens. The latter approach offers several advantages: it does not require derivative computations, meaning there is no need to access the model's internal parameters and gradients. This allows working with black-box models and avoids additional computational overhead \cite{liu2023pre}. Additionally, it preserves prompt interpretability, enabling humans to analyze and edit them \cite{liu2023pre}, and allows optimization for any metric (including non-differentiable ones) \citep{guo2023evoprompt, li2023spell, pan2023plum, fernando2023promptbreeder}.

However, this approach also has challenges: the optimization space for prompts is vast, and prompts generated through search methods may lack diversity \cite{guo2023evoprompt}. Nevertheless, there are numerous heuristic optimization algorithms that employ stochastic strategies, making the optimization process less sensitive to local optima. Evolutionary algorithms are one such example \cite{eiben2015evo}.

In this work, we analyzed the Reflective Evolution algorithm \cite{ye2024reevo} and integrated it to address the problem of automatic prompt generation. The resulting solution, called ReflectivePrompt, was tested on 33 datasets to demonstrate its effectiveness compared to existing methods.

\subsection{Evolutionary algorithms}

Evolutionary algorithms are a family of optimization methods based on the principles of biological evolution: natural selection, mutation, crossover, and inheritance. These algorithms operate with populations of solutions, gradually improving them according to a given fitness function [18]. Among such algorithms, the genetic algorithm \cite{holland1992ga} can be distinguished, which works with gene sequences (in our case, sequences of phrases in prompts). Within this algorithm, starting with an initial population of individuals, selection, crossover of selected individuals (creating offspring based on a combination of parental information), mutation of the offspring (random modification of certain parts), and population update based on offspring evaluations are performed iteratively.

This approach is highly flexible when applied to problems from various domains. The usage of evolutionary operators (crossover, mutation) and population-based search reduces the risk of getting stuck in local optima, maintaining a balance between exploring new solutions and exploiting existing ones, thereby leading to a high diversity of individuals in the final population while ensuring their quality according to the objective function remains high \citep{guo2023evoprompt, li2023spell, pan2023plum, fernando2023promptbreeder}.

\subsection{Related works}

One solution employing genetic algorithms is EvoPrompt \cite{guo2023evoprompt}. The improvement of the candidate prompt population occurs iteratively through selection, evolution (generation of new candidates using evolutionary operators), and population updates based on the evaluation of new candidates. The implementation of evolutionary operators (mutation and crossover) is achieved through queries to an LLM, enabling the utilization of its expertise in solving NLP tasks while maintaining prompt readability. During new candidate generation, two parents are first selected from the previous population using roulette-wheel selection (selection phase) \cite{lipowski2011roulette}, followed by the application of crossover and subsequent mutation of the resulting offspring. The study also presents a differential evolution algorithm \cite{storn1997de}, which involves mutating different segments of two donor prompts from the same population, combining them with a mutating candidate, and performing crossover with the current best prompt. The authors’ position EvoPrompt as a general framework for integrating LLMs into evolutionary algorithms, with experimental results demonstrating that differential evolution exhibits superior performance on more complex tasks.

SPELL \cite{li2023spell} employs a genetic algorithm operating iteratively through repeated reproduction and selection steps, where selection is performed via roulette-wheel while reproduction involves generating offspring based on a list of parent prompts and their corresponding scores. Notably, although reproduction is also conducted through LLM queries, this solution lacks explicit separation between crossover and mutation. Instead, it utilizes a predefined prompt instructing modifications to the parent prompt set (replacing, adding, or deleting words, altering tone) to generate offspring.

An alternative approach implemented in Plum \cite{pan2023plum} is based on metaheuristics. Unlike EvoPrompt's prompt mutation methodology, Plum explicitly defines a set of prompt modification operations: adding, deleting, rephrasing words/phrases, or swapping their positions, thereby generating multiple neighboring prompts. The solution architecture comprises: a well-defined set of neighboring prompts for each prompt, a metaheuristic algorithm with its inherent hyperparameters, and auxiliary functions (including crossover). The study examines six algorithms: hill climbing \cite{russel2016ai}, simulated annealing \cite{kirkpatrick1983simulatedann}, genetic algorithm (two variants - with mutation and crossover, and mutation-only) \cite{holland1992ga}, tabu search \cite{glover1986ai}, and harmony search \cite{geem2001harmony}. Each algorithm performs candidate mutation through the application of predefined modification operations, enabling exploration within the discrete prompt space. It should be noted that only the rephrasing operation is executed via LLM queries, while other operations are performed manually. As described by the authors, experimental results demonstrate this approach's capability to identify novel structural prompt modifications that enhance performance.

In Promptbreeder \cite{fernando2023promptbreeder} paper, the authors propose an extended genetic algorithm mutation approach incorporating predefined mutation prompts and "thinking styles" (concise descriptions of cognitive strategies, e.g., "Let's think step by step"), in addition to utilizing Chain-of-Thought \cite{wei2022chainofthought} and Plan-and-Solve \cite{wang2023planandsolve} techniques. At each iteration, candidates are improved through the application of a randomly selected mutation from a uniform distribution. The authors identify five mutation classes: direct mutation, hypermutation, estimation of distribution mutation \cite{larranaga2002distribution}, Lamarckian mutation \cite{ross1998lamarckian}, and prompt crossover/context shuffling. The first two classes further include zero-order and first-order mutations, totaling ten distinct mutations, each implemented through LLM queries. First-order direct mutation modifies candidates using specific mutation prompts, while zero-order mutation utilizes the initial problem statement to address method divergence. When problem specifications lack precision, Lamarckian mutation facilitates prompt reconstruction based on the last output yielding correct results. The algorithm's key innovation involves hypermutation, which modifies the mutation prompts themselves (via hypermutation prompts), thereby enhancing not only prompt solutions but also the improvement mechanisms. According to the authors, this diversity of operators enables continuous reformulation and representation of problems by LLMs, leading to more effective solutions \cite{fernando2023promptbreeder}. This approach demonstrates adaptability across various domains while optimizing prompts and preserving their interpretability.

\section{ReflectivePrompt}

\subsection{Reflective Evolution}

Reflective evolution is an approach described in the article ReEvo: Large Language Models as Hyper-Heuristics with Reflective Evolution \cite{ye2024reevo}. Its essence lies in using a language model to generate prompts aimed at enhancing the efficiency of mutation and crossover operations. The processes of prompt creation are referred to as short-term and long-term reflection. According to the authors, such reflective actions can be interpreted as obtaining a "verbal gradient" within the prompt space. Short-term reflection involves generating crossover prompts based solely on the current parent population, while long-term reflection, as the name suggests, entails accumulating knowledge, dependencies, and methods for improving efficiency throughout the entire evolutionary operation.

The application of reflection helps guide the direction of mutation and crossover operations while also expanding the search space, potentially moving beyond the initial population's predefined prompt space. In the original article, this approach was successfully applied to solving problems such as Guided Local Search (GLS) \cite{voudouris2010gls}, Ant Colony Optimization (ACO) \cite{dorigo1996aco}, Electronic Design Automation (EDA) \cite{shibasaka2013edaps}, the Decap Placement Problem (DPP) \cite{kim2022cadp}, the Traveling Salesman Problem (TSP) \cite{gohil2022tsp}, and other combinatorial optimization tasks. 

\subsection{Proposed solution}

In this study we developed a novel approach that combines methods of reflective evolution with large language models for the automatic generation of higher-quality prompts — ReflectivePrompt. ReflectivePrompt employs short-term and long-term reflection operations for subsequent use in crossover and elitist mutation. All performed operations and their corresponding queries to the language model were modified and refined to directly optimize prompts.

Specifically, the beginning of each instruction was changed to: "You are an expert in the domain of optimization prompts. Your task is to give hints to design better prompts." This adjustment is motivated by the specifics of the autoprompting task, for which reflective evolution was applied. Techniques describing possible modification operations performed during crossover and mutation, previously used in the SPELL algorithm, were incorporated. Thus, the following was added to the model queries defining short-term and long-term reflection: "For example, you can try to recommend word replacements, active/positive voice conversions, adding words, or deleting words." This enables the LLM to generate more precise, well-described hints that affect not only the semantic content of prompts but also their structural aspects.

ReflectivePrompt simplifies user interaction by generating an initial population of prompts based on just a single input prompt. In this approach, the prompt is rephrased using the LLM and structured output techniques \cite{liu2024stuctured}.

A key feature of ReflectivePrompt is delegating the decision on the specifics of mutation to the model itself. In previously described solutions, the mutation type was either predefined and fixed or randomly selected from a uniform distribution. In this approach, however, the model generates hints autonomously and tends to decide whether to apply structural transformations to the prompt or only modify its semantic meaning and phrasing.

When performing crossover and mutation operations, the LLM is provided with a brief task description, which helps generate more problem-targeted prompts while preserving the logical structure of the instruction. Empirical observations have shown that even large models can achieve decent metric values using prompts that are partially or entirely irrelevant to the task. As a result, the final prompts may deviate significantly from the intended meaning. ReflectivePrompt avoids this issue and, in the vast majority of cases, generates semantically correct prompts that are more comprehensible to human perception and logic.	

\begin{figure}[H]
    \centering
    \includegraphics[width=1\textwidth]{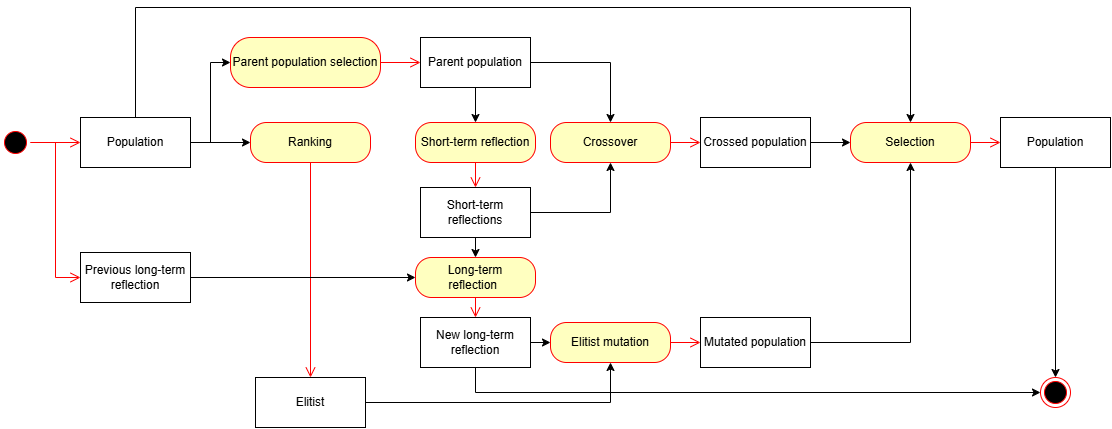}
    \caption{The Reflective Evolution pipeline in ReflectivePrompt}
    \label{fig:reflectivepipeline}
\end{figure}

The general scheme of reflective evolution within ReflectivePrompt is illustrated in Figure \ref{fig:reflectivepipeline}.

Particular attention should be paid to the two selection operations. The parent population selection of prompts chooses pairs of parent prompts from the current population. In this process, each prompt can be included in multiple parent pairs. The main constraint, which is related to the original reflective evolution algorithm, is that prompts in a parent pair must have different fitness function values. Parent selection is performed using the roulette-wheel method \cite{lipowski2011roulette}. The probability vector for being selected for each individual is represented by the normalized vector of their fitness scores. The second selection operation, which mimics the survival of the fittest, also employs the roulette-wheel method, but in this case, the probabilities are obtained by applying a softmax operation with a temperature of 0.1 to the fitness function value vector. This temperature value yields a less uniform distribution in cases where all prompts have approximately similar scores, thereby increasing the probability of selection for individuals with higher fitness values.

Another crucial aspect is the preservation of elite individuals in the population. Before the start of each epoch, the individual that has demonstrated the best performance throughout the entire evolutionary process is reintroduced into the population, even if it was not selected at the end of the previous iteration. This approach enhances the algorithm's convergence speed, as the best individuals are not lost over time due to unfavorable selection outcomes.

\begin{figure}[H]
	\centering
        \includegraphics[width=0.6\textwidth]{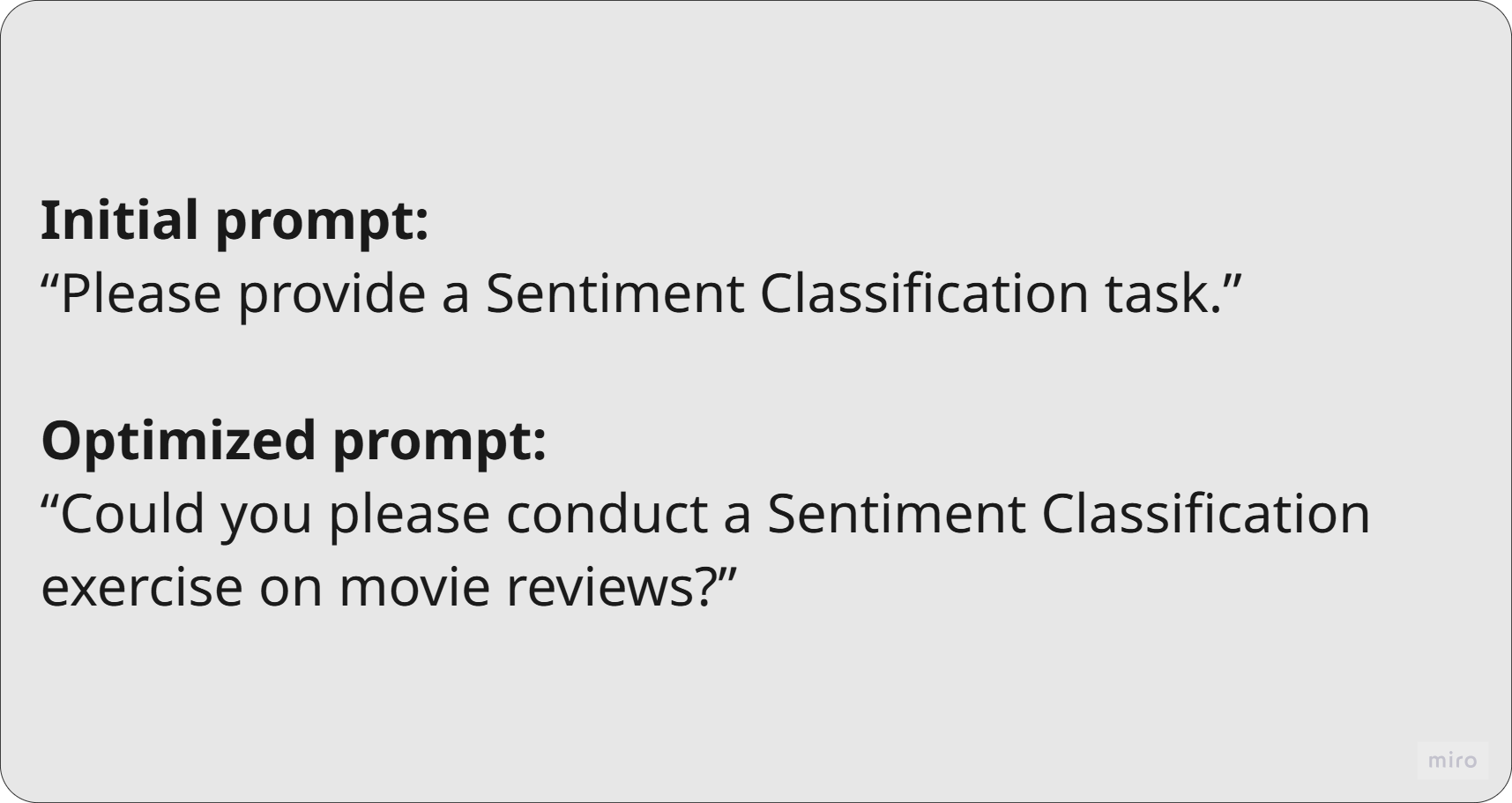}
	\caption{The optimized prompt for SST-2 dataset}
	\label{fig:sst2_example}
\end{figure}

\begin{figure}[H]
	\centering
        \includegraphics[width=1\textwidth]{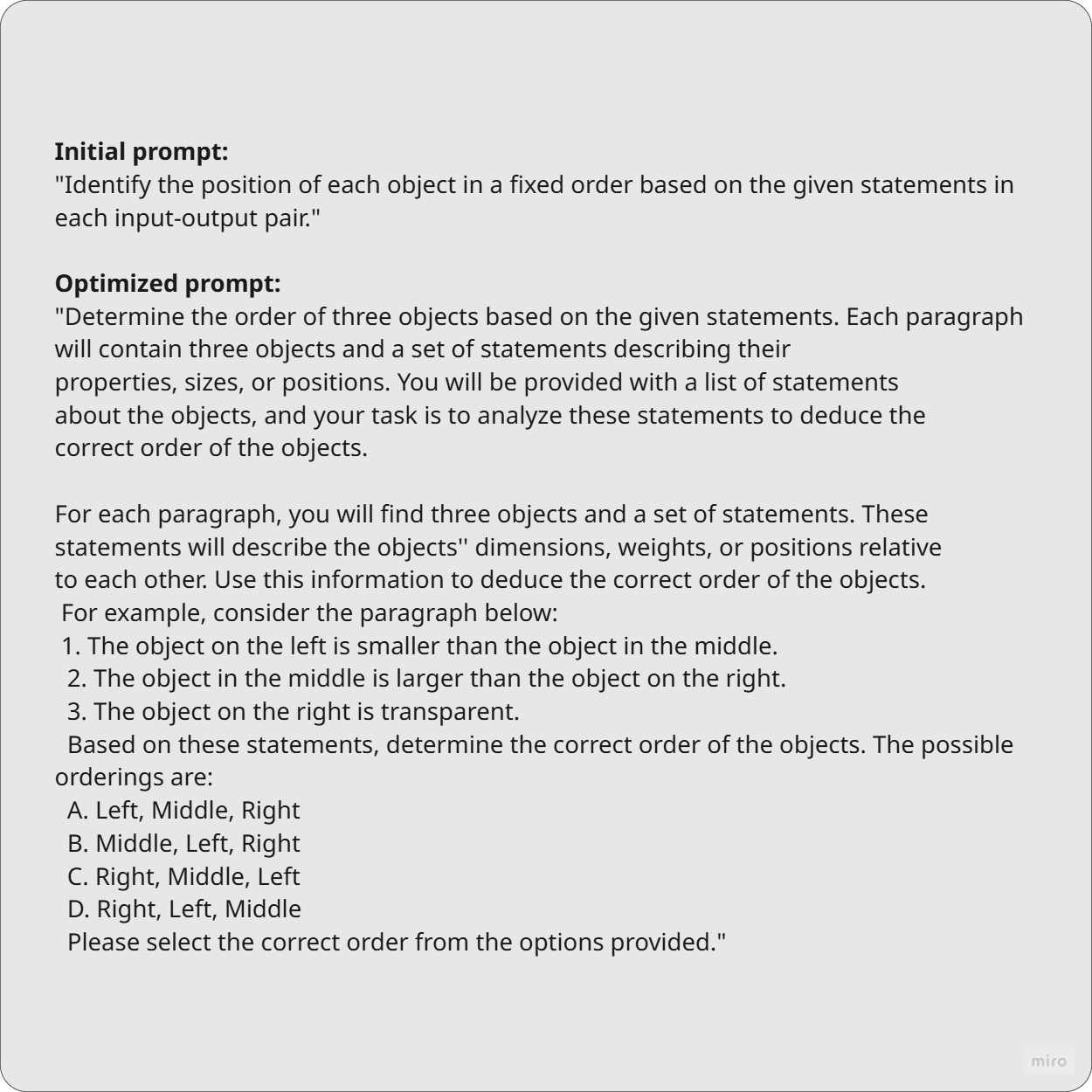}
	\caption{The optimized prompt for BBH/logical\_deduction\_three\_objects}
	\label{fig:bbhexample}
\end{figure}

The examples of ReflectivePrompt optimization are shown in Figures \ref{fig:sst2_example} and \ref{fig:bbhexample}.

\section{Experimental Evaluation}
\label{sec:evaluation}

\subsection{Experimental Setup}
\label{subsec:exp_setup}

ReflectivePrompt was evaluated on 33 datasets for text classification and generation tasks. As baselines and reference points for comparison, we used results from EvoPrompt, SPELL, PromptBreeder, and Plum. The autoprompting algorithms were executed using large language models from different families and sizes (t-lite-instruct-0.1, gemma3-27b-it \cite{kamath2025gemma}). This choice of LLMs was made due to the use of open source white-box models which are more user-friendly and can be utilized by everyone. Also the significant difference in the number of model parameters leads to better testing coverage and makes our results more unbiased.

\subsection{Classification tasks}

\begin{figure}[H]
	\centering
        \includegraphics[width=1\textwidth]{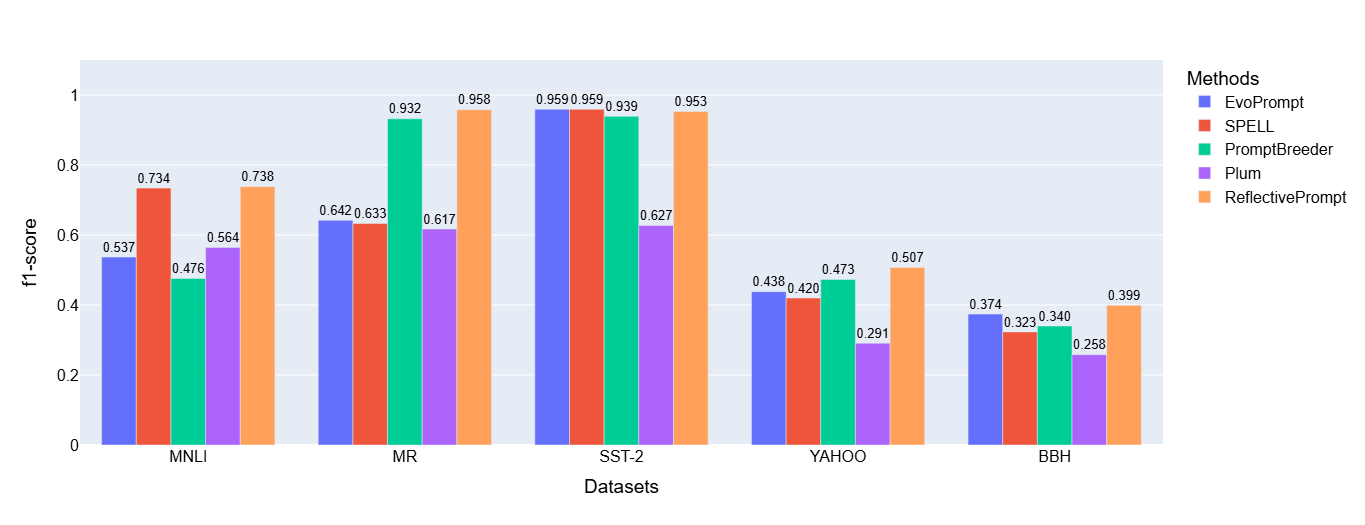}
	\caption{Histogram of F1-score values. Model: t-lite-instruct-0.1}
	\label{fig:f1tlite}
\end{figure}

\begin{figure}[H]
	\centering
        \includegraphics[width=1\textwidth]{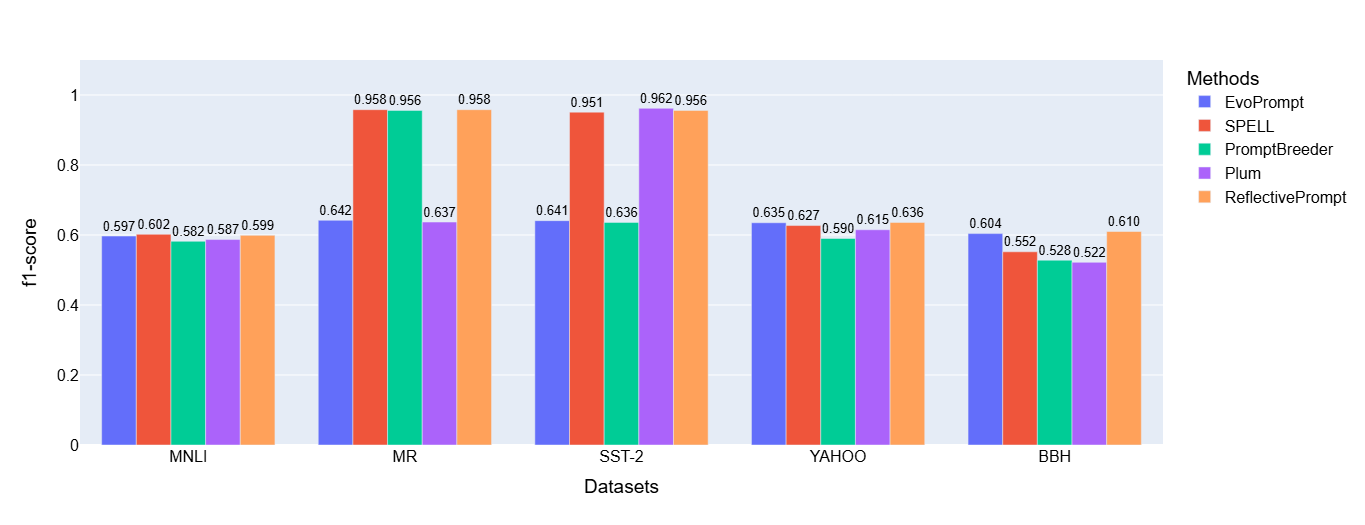}
	\caption{Histogram of F1-score values. Model: gemma3-27b-it}
	\label{fig:f1gemma}
\end{figure}

For classification tasks, the following datasets and benchmarks were used: MNLI, MR, SST-2, YAHOO, and BBH (a subset of datasets with strictly formatted answers that can be treated as classification tasks). The metric selected for evaluation and optimization during evolution was the F1-score. The results of each method are presented in Figures \ref{fig:f1tlite}-\ref{fig:f1gemma}.

\subsection{Generation tasks}

\begin{figure}[H]
	\centering
        \includegraphics[width=1\textwidth]{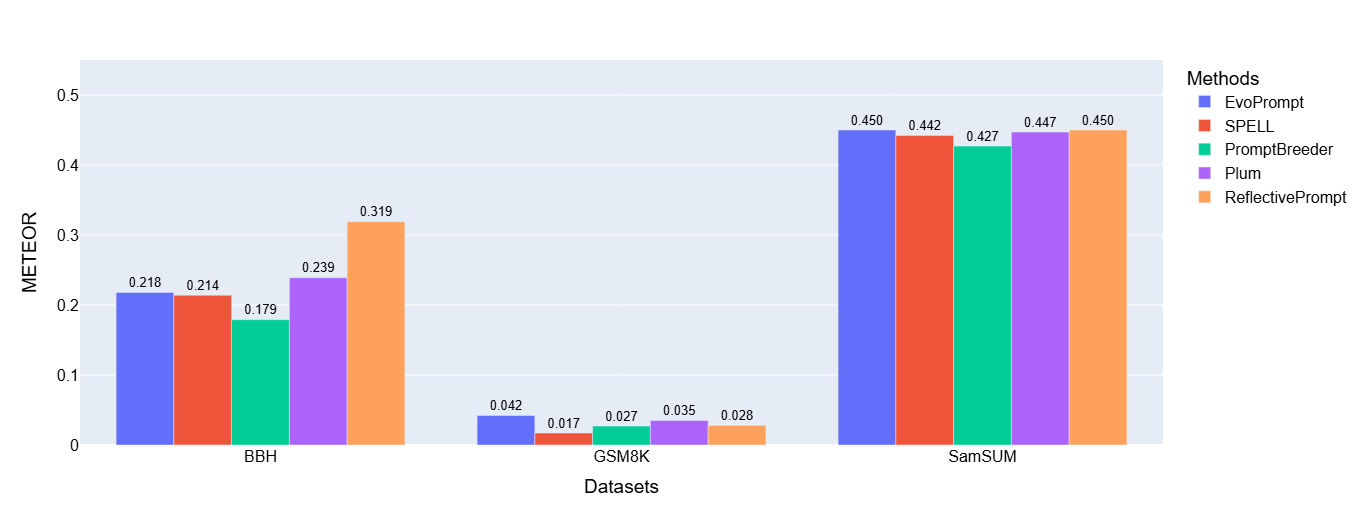}
	\caption{Histogram of METEOR scores for text generation datasets. Model: t-lite-instruct-0.1}
	\label{fig:meteortlite}
\end{figure}

\begin{figure}[H]
	\centering
        \includegraphics[width=1\textwidth]{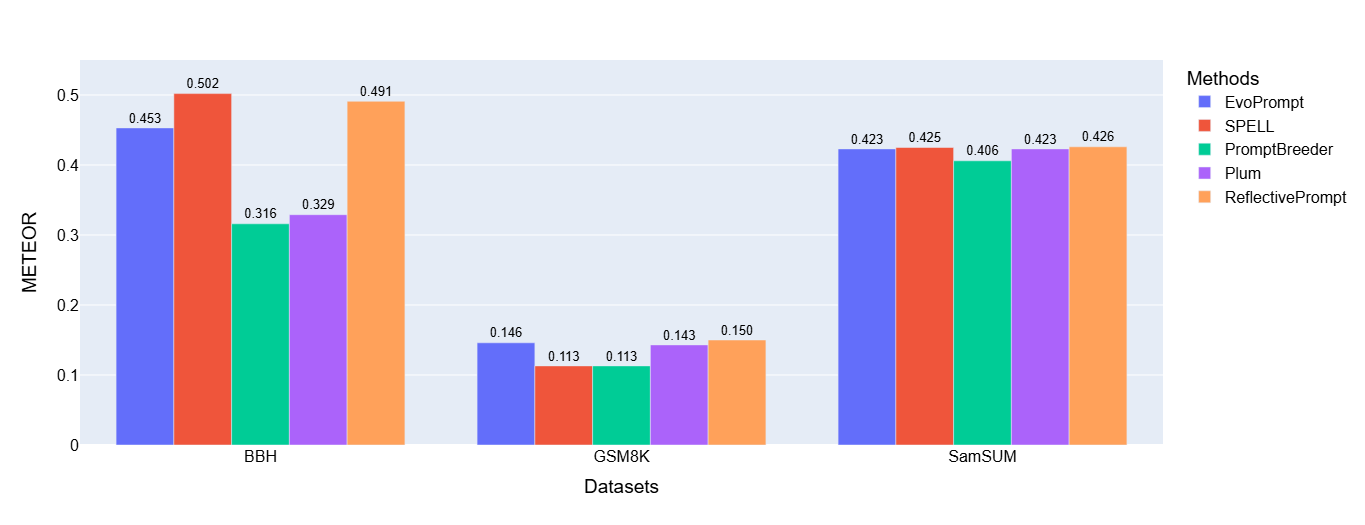}
	\caption{Histogram of METEOR scores for text generation datasets. Model: gemma3-27b-it}
	\label{fig:meteorgemma}
\end{figure}

ReflectivePrompt and its counterparts were evaluated on the following datasets: BBH (dyck\_languages, multistep\_arithmetic\_two, object\_counting, word\_sorting), GSM8K, and SamSUM. The metric used for evaluation and optimization was METEOR. The main results are shown in Figures \ref{fig:meteortlite}-\ref{fig:meteorgemma}.

\section{Discussion}
\label{sec:discussion}

The conducted experiments demonstrate that ReflectivePrompt effectively handles both classification and text generation tasks. Across all evaluated datasets, ReflectivePrompt either outperformed or matched the performance of existing evolutionary algorithm-based autoprompting methods. The method showed particularly strong results on the BBH benchmark, comprising 23 classification tasks and 4 text generation tasks. For classification tasks, the average F1-score improved by 6.59\% on the t-lite-instruct-0.1 model and by 0.96\% on the gemma3-27b-it model. In text generation tasks, the average METEOR score increased by 33.34\% on the t-lite-instruct-0.1 model (comparisons and improvements were calculated relative to the maximum average metrics achieved by existing solutions).
It should be noted that ReflectivePrompt's performance significantly depends on the underlying LLM. The effectiveness of reflective evolution relies on the quality of generated hints, and weaker language models may produce suggestions that are not fully relevant to the optimization task.
This work creates a scope for future research into reflective evolution for autoprompting applications. The current ReflectivePrompt implementation could potentially be further refined for more targeted prompt optimization. Moreover, the concept of reflective evolution could be generalized and adapted to other metaheuristic optimization algorithms, representing a promising direction for future studies. For example, there was a recent research where reflective prompt evolution outperforms reinforcement learning on a group of benchmarks \cite{agrawal2025gepa}.

\section{Conclusion}
\label{sec:conclusion}

The proposed ReflectivePrompt algorithm, which employs reflective evolution for prompt optimization, was evaluated on 33 datasets covering various natural language processing domains. It demonstrated consistent improvements over existing evolutionary algorithm-based autoprompting methods. ReflectivePrompt proves to be a competitive solution, showing that exploring reflective evolution for autoprompting can yield significant benefits and advance current methods to new levels of performance.

\bibliographystyle{plain}
\bibliography{references}

\end{document}